\documentclass{article}

\usepackage{microtype}
\usepackage{graphicx}
\usepackage{subfigure}
\usepackage{booktabs} 
\usepackage{amsmath}
\usepackage{amssymb}
\usepackage{bbding}
\usepackage{hyperref}

\usepackage[accepted]{icml2021}

\usepackage{xspace}
\usepackage{balance}
\usepackage{pifont}

\newcommand{\subj}{FedHPO\xspace}
\newcommand{\psubj}{pFedHPO\xspace}

\begin{document}

\twocolumn[
\icmltitle{HPN: Personalized Federated Hyperparameter Optimization}




\begin{icmlauthorlist}
\icmlauthor{Anda Cheng}{to,in}
\icmlauthor{Zhen Wang}{goo}
\icmlauthor{Yaliang Li}{goo}
\icmlauthor{Jian Cheng}{to}
\end{icmlauthorlist}

\icmlaffiliation{to}{Institute of Automation, Chinese Academy of Sciences.}
\icmlaffiliation{goo}{Alibaba Group}
\icmlaffiliation{in}{The work was done when A. Cheng was an intern at Alibaba Group.}

\icmlcorrespondingauthor{Zhen Wang}{jones.wz@alibaba-inc.com}
\icmlcorrespondingauthor{Yaliang Li}{yaliang.li@alibaba-inc.com}


\vskip 0.3in
]



\printAffiliationsAndNotice{}  

\begin{abstract}
Numerous research studies in the field of federated learning (FL) have attempted to use personalization to address the heterogeneity among clients, one of FL's most crucial and challenging problems. However, existing works predominantly focus on tailoring models.
Yet, due to the heterogeneity of clients, they may each require different choices of hyperparameters, which have not been studied so far. We pinpoint two challenges of personalized federated hyperparameter optimization (pFedHPO): handling the exponentially increased search space and characterizing each client without compromising its data privacy. To overcome them, we propose learning a \textsc{H}yper\textsc{P}arameter \textsc{N}etwork (HPN) fed with client encoding to decide personalized hyperparameters. The client encoding is calculated with a random projection-based procedure to protect each client's privacy. Besides, we design a novel mechanism to debias the low-fidelity function evaluation samples for learning HPN. We conduct extensive experiments on FL tasks from various domains, demonstrating the superiority of HPN.
\end{abstract}

\section{Introduction}
\label{sec:intro}
With increasing concerns about privacy issues in recent years, federated learning (FL)~\cite{deffl} has gained more and more attention from both industries and academia. Despite its various settings and diverse application scenarios, FL, in principle, aims to learn models collaboratively from dispersed data while keeping data privacy intact. To achieve this, FL participants often keep their data inaccessible to others while exchanging model-related information in the parameter or function space. An example is the popular FL algorithm FedAvg and its variants~\cite{fedoptimizationsurvey}, wherein the server optimizes the model by repetitively broadcasting the latest parameters and aggregating the local updates made by clients.

One of the most crucial factors that make FL distinct from general distributed machine learning is the heterogeneity among clients~\cite{fs}. In general distributed machine learning, workers have independently identically distributed (i.i.d.) data. In contrast, clients may have heterogeneous data (non-i.i.d.), and thus making aggregation from their local updates is non-trivial and has become one of the core research problems in the field of FL~\cite{flsurvey,flsurvey1}.

Among the endeavors of handling client heterogeneity, personalization is a promising and popular strategy, which has been instantiated as many successful personalized FL (pFL) algorithms~\cite{pflbench}. Conceptually, pFL algorithms allow each client to learn and apply a model tailored to its data distribution. Such client-wise models are often expected to improve both overall performance and fairness~\cite{ditto}.

Existing personalization methods mainly focus on the model but overlook other aspects that also affect the ultimate performance, such as hyperparameters. Intuitively, the data heterogeneity may lead to different optimal hyperparameter configurations for the clients, e.g., a larger learning rate is in favor if the absolute value of input features is smaller than other clients. Moreover, despite the statistical heterogeneity, there has been a surge of federated hetero-task learning~\cite{hete-task}, where FL participants may have different learning tasks of varying feature and label space. We identify and verify that clients indeed have respective optimal choices of the hyperparameters on a federated hetero-task learning dataset~\cite{fedhpobench} (see Sec.~\ref{sec:exp}). These observations motivate us to study personalized federated hyperparameter optimization (pFedHPO).

However, it is challenging to achieve personalization for HPO. At first, a pFedHPO algorithm must be sample efficient since, in essence, the search space has been extended exponentially along with the number of clients. Moreover, FL requires communications among multiple participants, making the evaluation of hyperparameter configuration extremely expensive. Hence, FedHPO is often conducted with a very restricted number of trials, under which circumstance it becomes more challenging to tackle such an enormous search space. Second, another desideratum of a pFedHPO algorithm is the capacity to characterize each FL participant’s data distribution while reducing the privacy leakage risk. Specifically, how to personally choose the appropriate hyperparameter configuration must depend on the specific data distribution. However, information related to each client’s data distribution should be carefully protected.

To overcome these challenges, we propose HPN, a novel method to achieve pFedHPO. Instead of a context-free bandit model widely adopted in existing FedHPO works (e.g., FedEx~\cite{fedex}), we design a policy network that takes a client encoding as input and outputs a distribution over the original search space. Therefore, each client has its specific encoding to determine its hyperparameters personally. Meanwhile, this policy network's parameters are updated based on the observations collected from all the clients, which can be regarded as improving the sample efficiency via model sharing. In HPN, a client encoding is calculated based on the training sample of that client to reflect the similarity among clients. For the purpose of privacy protection, we leverage Random Fourier Feature (RFF)~\cite{rff} to transform the extracted features to reduce the risk of privacy leakage. Furthermore, we design a mechanism to conduct low-fidelity evaluations of hyperparameter configurations, which reduces variance for the signals used to update our policy and, in the meantime, alleviates the impact of model parameters' state. Finally, we conduct extensive experiments on FedHPO tasks of various domains, where the generalization error of the hyperparameter configurations searched by HPN is lower than that of related baselines.

Our contributions are summarized as follows:
\begin{itemize}
	\item We are the first to systematically explore pFedHPO, discussing its setting, challenges, and critical problem-solving factors.
	
	\item We propose a novel pFedHPO method HPN, which satisfies the sample efficiency and privacy preservation requirements. Meanwhile, we design a mechanism to resolve an issue in sampling low-fidelity function evaluations that prior works have ignored.
	
	\item Extensive experiments on multiple FL datasets of various modalities suggest the benefit of pFedHPO and show the effectiveness of our proposed HPN
	
\end{itemize}

\section{Federated Hyperparameter Optimization and Its Personalized Setting}
\label{sec:pre}
In this paper, we largely follow the problem setting and notations considered in FedEx~\cite{fedex}, yet we extend \subj into a personalized setting, which brings in more challenges.

\noindent\textbf{FedHPO}.
Despite the various settings and corresponding algorithms of FL, we focus on a lens of popular FL algorithms, including FedAvg~\cite{deffl}, FedOpt~\cite{fedopt}, FedProx~\cite{fedprox}, etc. Such algorithms can be categorized into two phases, one for \textit{aggregation} and the other for \textit{local training}. Thus, we denote the hyperparameter search space by $\mathcal{A}=\mathcal{B}\times\mathcal{C}$, where $\mathcal{B}$ and $\mathcal{C}$ represent the subspace corresponding to \textit{aggregation} and \textit{local training}, respectively. Suppose there are $n$ heterogeneous clients composing a federation, each of which has the training, validation, and testing dataset $T_i$, $V_i$, and $E_i$, respectively. Then \subj can be formulated as solving the following optimization problem:
\begin{equation}
	\begin{aligned}
		\label{eq:originaldef}
		&\min_{a=(b,c)\in\mathcal{A}}\sum_{i=1}^{n}|V_i|\mathcal{L}_{V_i}(\mathbf{w}_{a}^{*})\\
		\text{s.t.}\quad&\mathbf{w}_{a}^{*}=\operatorname{Alg}_a(\{T_{j}|j=1,\ldots,n\}),
	\end{aligned}
\end{equation}
where the algorithm is configured by $a$ and federally learn a model parameterized by $\mathbf{w}_{a}^{*}\in\Re^d$ from all the training datasets, the learned model parameters $\mathbf{w}_{a}^{*}$ is evaluated on all the validation datasets, and the cardinality of validation dataset weights the resulting validation losses to form the objective.

\noindent\textbf{Approximation}.
As executing a complete FL course is costly, multi-fidelity HPO methods are highly in demand for FL, where training the model for fewer communication rounds before evaluating its validation performance is one of the most popular approximations~\cite{fedhpobench}. However, traditional multi-fidelity methods, such as Successive Halving Algorithm (SHA)~\cite{sha}, make hard elimination of candidate configurations, which is troublesome due to the noisy validation performance federally attained from just a portion of all validation datasets~\cite{fedex}.

To alleviate this issue, FedEx~\cite{fedex} restricts its scope to optimizing only the local training-related hyperparameters and incorporates the weight-sharing trick of neural architecture search (NAS)~\cite{enas} into \subj, where the following single-level surrogate problem is to be solved instead of the original bi-level one (i.e., Eq.~\eqref{eq:originaldef}):
\begin{equation}
	\label{eq:originalpsndef}
	\min_{c\in\mathcal{C},\mathbf{w}\in\Re^d}\sum_{i=1}^{n}|V_i|\mathcal{L}_{V_i}(\operatorname{Loc}_{c}(T_{i},\mathbf{w})),
\end{equation}
where the local training phase is configured by $c$ and returns updated model parameters resulted from updating $\mathbf{w}$ over $T_i$. Then both the model parameters and the decisions of hyperparameters could be simultaneously optimized.

\noindent\textbf{pFedHPO}.
As discussed in Sec.~\ref{sec:intro}, client heterogeneity may demand client-wise configurations, which essentially means an extended search space $\mathcal{A}=\mathcal{B}\times\mathcal{C}_{1}\times\cdots\mathcal{C}_{n}$.
Then we naturally define the \psubj as solving Eq.~\eqref{eq:originaldef} while replacing the original search space with this extended version.
It is worth mentioning that, in our formulation, the ultimate goal of \psubj is to seek client-wise configurations that optimize the FL algorithm of interest. In contrast, the personalized setting discussed in FedEx intends to seek one global configuration that optimizes the procedure of the original FL algorithm appended with a client-specific fine-tuning phase.

\begin{figure*}[tbhp]
	\centering
	\label{fig:framework}
	\includegraphics[width=.93\linewidth]{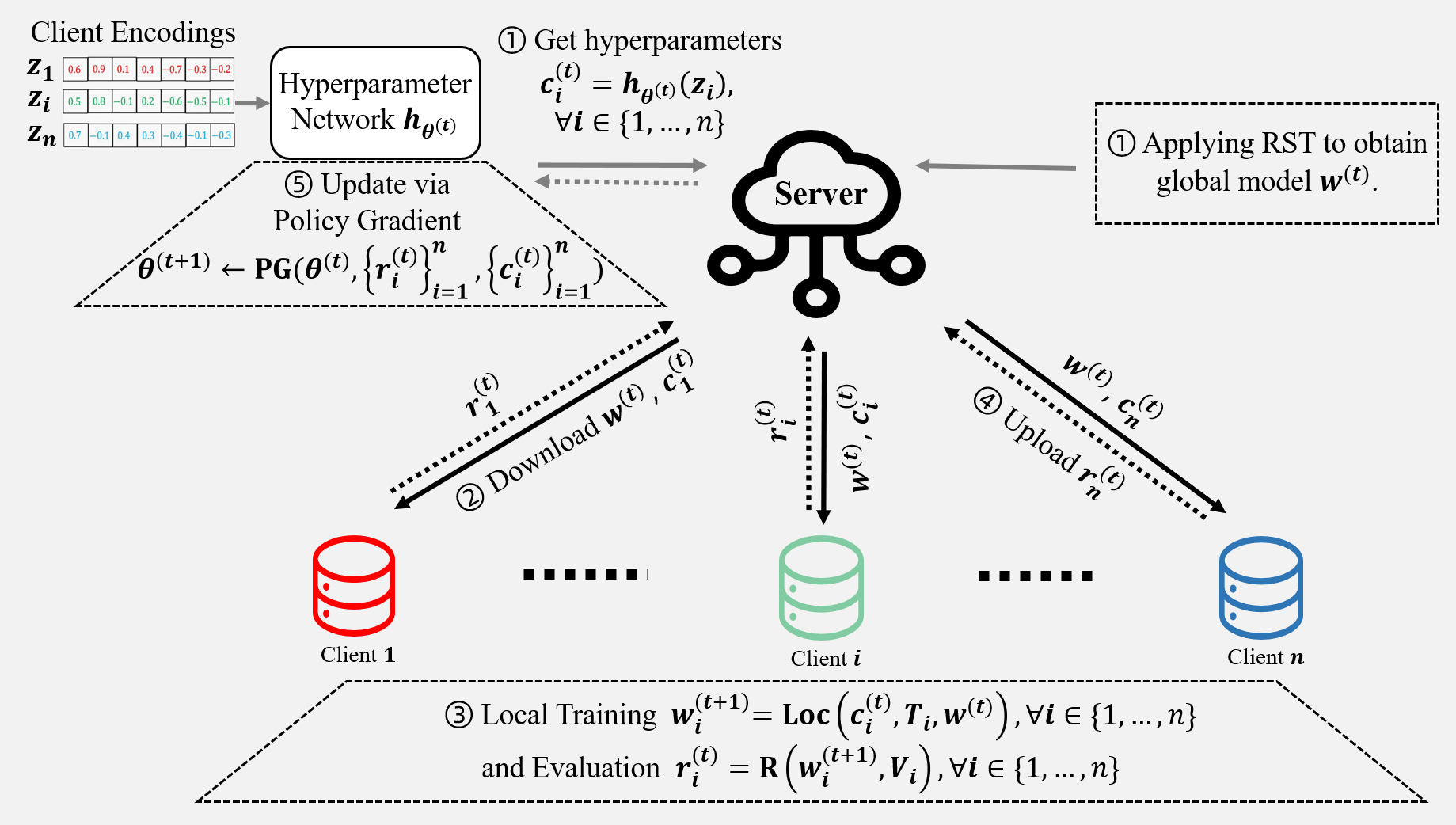}
	\caption{HPN learning framework.
		At each learning step, 
		\ding{172} the server first takes client encodings as input and outputs personalized configurations for different clients using HPN and applies RST (Sec.~\ref{subsec:rst}) to obtain global FL model for this trial.
		\ding{173} the server broadcasts the global FL model and configurations to clients.
		\ding{174} clients run local training and evaluation using received global model and configurations to get new local models and validation results.
		\ding{175} clients return local models and the corresponding evaluation results to the server.
		\ding{176} the server updates HPN using the received evaluation results via policy gradient method.
	}
\end{figure*}

\noindent\textbf{Challenges}.
This extended search space increases exponentially with $n$, which, intuitively speaking, needs much more trials to seek the optimal configuration than the original \subj problem. Considering the cost of each trial, namely, an FL training course consisting of tens to hundreds of communication rounds, applying a classic Bayesian optimization method to solve a \psubj problem becomes impractical.
In addition to this enormous search space, how to characterize each client's data distribution without compromising data privacy is another challenge.

\noindent\textbf{Problem Statement}.
As what is personalized is the local training-related hyperparameters, we also restrict our goal to tuning them. Meanwhile, we also consider the single-level formulation proposed by FedEx and adopt the stochastic relaxation of differentiable NAS~\cite{DARTS}:
\begin{equation}
	\label{eq:redefinedpfedhpo}
	\min_{\theta\in\Theta,\mathbf{w}\in\Re^d}\sum_{i=1}^{n}|V_i|\mathbb{E}_{c_{i}\sim\pi_{\theta}}[\mathcal{L}_{V_i}(\operatorname{Loc}_{c_i}(T_{i},\mathbf{w}))],
\end{equation}
where client-wise configurations $c_{i},i=1,\ldots,n$ are drawn from a policy $\pi$ parameterized by $\theta$. We identify two critical factors for optimizing this objective:
\begin{itemize}
	\item How to model $\pi_\theta$? If we trivially consider $\mathcal{C}_{1}\times\cdots\times\mathcal{C}_{n}$ as its action space and let $\pi_\theta$ be client-agnostic, then the policy learning would pose an unacceptable sample complexity. If we would otherwise consider $\mathcal{C}$ as the action space and an input context to reflect the characteristics of a client, the second challenge identified above has to be overcome.
	\item How to schedule the evaluations of hyperparameter configurations regarding stages of an FL course? In both differentiable NAS and FedEx, $\theta$ is updated based on the collected validation performances no matter which stage of a training course currently is, namely, ignoring the state of $\mathbf{w}$. Then most related methods, such as FedEx, gradually collects the validation performances along with the training course of $\mathbf{w}$. Such a schedule implies a strong preference towards hyperparameter configurations that is advantageous in the early stage.
\end{itemize}

\section{Our Method}
\label{sec:method}
Considering the above two factors in optimizing Eq.~\eqref{eq:redefinedpfedhpo}, we propose to learn a HyperParameter Network (HPN) shared by all the clients to decide their hyperparameter configurations based on their client encodings. This sharing and, thus, the search space $\mathcal{C}$ kept the same as the non-personalized setting are beneficial for reducing the sample complexity. To learn HPN better, we also design a mechanism named random start training to alleviate the influence of the state of model parameters. The framework of our method is shown in Fig.~\ref{fig:framework}. Then, we elaborate on HPN, how we calculate client encoding, and the random start training mechanism.

\subsection{Hyperparameter Network for pFedHPO}\label{subsec:hpn}

Personalized hyperparameter optimization in federated learning is highly challenging due to the exponential growth of the search space with the number of clients and 
the expensive computation and communication cost for HPO evaluation in FL.
To overcome these challenges, we propose to generate personalized hyperparameters via a hyperparameter network (HPN) that takes different client encoding vectors as input and outputs different distributions in the original search space.
Specifically, suppose we have a set of client encodings $\{z_i \}_{i=1}^n$ which are obtained via some encoding method based on each FL participant's data distribution, i.e. $z_i \leftarrow \operatorname{Enc}(T_i)$.
Our hyperparameter network $h_{\theta}$  takes client encodings as input and outputs parameters for a parameterized distribution. 
Then the personalized hyperparameters for different clients are sampled from the different parameterized distributions 
$c_i \sim \mathcal{P}\left( c \mid h_{\theta} \left( z_i \right) \right)$. 
Applying with the sampled hyperparameters $\{c_i \}_{i=1}^n$, the clients can collaboratively run a federated learning course (e.g. running FedAvg algorithm) on their training datasets $\{T_i \}_{i=1}^n$ to obtain a well-trained FL model $\mathbf{w}_{*}$.

The hyperparameter network $h_{\theta}$ can be trained by solving the following optimization problem:
\begin{align} \label{hpn}
	\min _{\theta \in \Theta, \mathbf{w} \in \mathbb{R}^{d}} \sum_{i=1}^{n}\left|V_{i}\right| \mathcal{L}_{V_{i}}\left(\operatorname{Loc}\left(h_{\theta}(z_i), T_{i}, \mathbf{w}\right)\right)
\end{align}
We can apply policy gradient method \cite{policy_grad} to solve this problem. 
Specifically, at each learning step $t$, the hyperparameter network first generates configurations for different clients to let them federally train a FL model.
Then the hyperparameter network is updated using policy gradient method by taking the performance of FL model on the validation dataset as the reward ${R}^{(t)}= -\sum_{i=1}^n \frac{|V_i|}{|V|} \mathcal{L}(\mathbf{w}_*^{(t)}, V_i)$ , which can be formally expressed as 
\begin{align}  
	\theta^{(t+1)} \leftarrow \theta^{(t)} + \alpha {R}^{(t)} \nabla_{\theta} \sum_{i=1}^n \log \mathcal{P}\left(c_i^{(t)} \mid h_{\theta^{(t)}}(z_i)\right)
\end{align}
where $\alpha$ is learning rate.
Repeating the above steps until convergence, we can obtain the final hyperparameter network $h_{\theta^*}$ which can output  hyperparameters for all clients according to their encoding vectors. 
Running a FL course with these hyperparameters will lead to a global FL model that achieves the minimum loss on the validation set. 

The hyperparameter distribution $\mathcal{P}$ can be either discrete or continuous, depending on the continuity of the search space. 
When the search space $\mathcal{C}$ is discrete, for each hyperparameter, there is a finite set of available choices. Consequently, $\mathcal{C}$ becomes a grid composed of all the possible combinations of searchable hyperparameters. 
In this case, we design the hyperparameter network as a multi-head network, with each output head corresponding to one hyperparameter and the output dimension equals to the number of candidate values for that hyperparameter in the search space. 
We select the discrete Categorical distribution as the prior distribution $\mathcal{P}$.
Each Categorical is then parameterized by a output vector from the head corresponding to a hyperparameter. 
Then we can sample hyperparameters according to the parameterized distributions for running FL course and updating hypernetwork.

When the search space $\mathcal{C}$ is continuous, we select a multi-variate Gaussian distribution as the prior distribution $\mathcal{P}$. 
In this case, each head of the hyperparameter network is two dimensional, with two dimensions corresponding to the mean and variance of the Gaussian distribution. 
By using the output of the hyperparameter network as the parameters for $\mathcal{P}$, we can then directly sample from the continuous multi-variate Gaussian distribution using the Box-Muller transformation \cite{box-muller}. The sampled values can then be mapped onto the hyperparameter space $\mathcal{C}$ to obtain the sampled hyperparameters.

\begin{figure*}
	\centering
	\includegraphics[width=1.\linewidth]{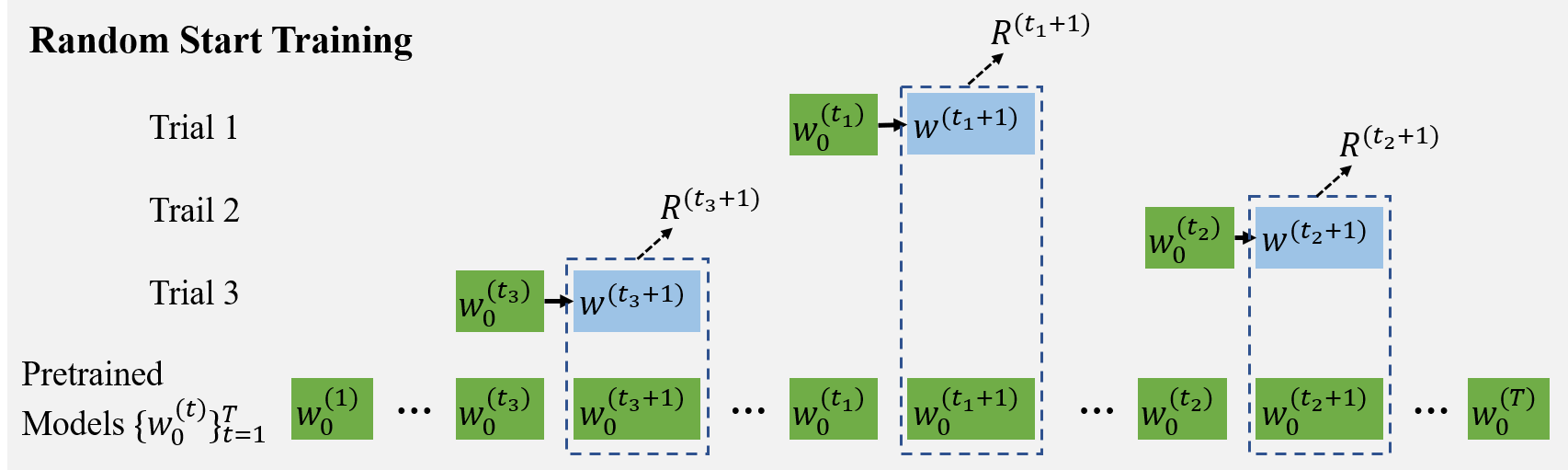}
	\vspace{-0.5cm}
	\caption{Random start training process.
		Before training HPN, RST firstly trains a complete FL course to save the global model checkpoints at each communication round.
		During the HPN training process, RST first samples a set of the hyperparameters and deploy them to clients, then randomly samples a starting round and loads the corresponding checkpoint.
		Starting from this model, RST runs the FL algorithm for a few communication rounds to get a new global model. Finally, we use the validation performance gain of the new global model over the saved model as reward to update HPN.}
	\label{fig:rst}
\end{figure*}

\subsection{Random Fourier Feature for Client Encoding}\label{subsec:rff}

The design of client encoding is non-trivial, since the client encoding vectors $\{ z_i\}^n_{i=1}$ must depend on and reflect
the clients' local data distribution. At the same time, they should not increase the risk of privacy violation compared to passing the raw
data.
To overcome this challenges, we propose to exploit random Fourier features (RFF) \cite{rff} for client encoding.
RFF are generated by randomly sampling frequency vectors from a normal distribution and then projecting data points onto them. 
As the frequency vectors are sampled from a normal distribution, RFF can capture the general shape of the overall data distribution \cite{rff18}. 
It has also been shown to reduce privacy leakage risk when used in machine learning applications \cite{dp-merf, fts}.
Thanks to its obfuscation effect, it is harder for an adversary to identify patterns in the raw data. Additionally, by transforming the data into another domain, it becomes more difficult for an adversary to uncover the original data structure.

Formally, RFF can be used to approximate any translation invariant kernel, i.e., $k\left(x, x^{\prime}\right) \approx \phi(x)^{\top} \phi\left(x^{\prime}\right)$,  for some function  $k(x, x^{\prime})$.
According to Bochner's theorem \cite{boch},  
the kernel function can be written as  
\begin{align}
	k\left(x-x^{\prime}\right)
	=\int p(\omega) e^{i \omega^{\top}\left(x-x^{\prime}\right)} \mathrm{d} \omega
\end{align}
where  $p(\omega) \triangleq \mathcal{N}(0, \mathrm{I}_D)$ is $D$-dimension normal distribution.
The kernel function $k\left(x-x^{\prime}\right)$  can be approximated using Monte Carlo sampling method.
The vector of RFF is given by
${\phi}(x)=\left({\phi}_{1}(x), \ldots,{\phi}_{D}(x)\right)^{\top}$, 
where ${\phi}_{j}(x) =\sqrt{2/D} \cos\left(\omega_{j}^{\top}x\right)$.
To get the client encoding vector for a participant, we  first apply the above Fourier transform to each example in the local training set to transform the example to frequency domain. 
Then the client encoding vector for this participant is obtained by averaging all the RFF vectors.

\subsection{Random Start Training for HPN}\label{subsec:rst}

In the section \ref{subsec:hpn}, we introduce how to use policy gradient method to train HPN. 
At each step, we sample hyperparameters from HPN to run a complete FL course and then use the resulting validation performance as reward to update HPN.
However, this full-fidelity training method is very inefficient, because in a complete FL course, the server and multiple clients need to conduct many rounds (usually $> 100$ rounds) of communication and computation.
Therefor, if we use the performance of the complete training FL model to evaluate the sampled hyperparameters, the cost of evaluation will be very expensive. 
A simple solution to this problem is to use low-fidelity evaluation methods \cite{fedhpobench}, such as not training a complete FL course but only running for a few communication rounds (e.g. $<5$ rounds). 
This method can effectively reduce the number of FL communication rounds in one trial.
However, the evaluation results of this method is very inaccurate, since the results only reflect the effect of the hyperparameters on the early stage of FL training but not the effect on the whole FL course.
Therefore, the most important question in training HPN is how to better balance the trade-off between efficiency and accuracy of evaluation
While facing with an exponentially enormous search space.

To solve this problem, we propose Random Start Training (RST). 
Suppose there are totally $T$ communication rounds that in a FL course.
The process of RST is shown in Fig.\ref{fig:rst}.
Before training HPN, RST first requires to use a set of hand-set hyperparameters to train a complete FL course (for $T$ rounds) and save the global model obtained after each communication round as $\{ \mathbf{w}_0^{(t)} \}^T_{t=1}$. 
At each step of HPN training, we first sample a set of the hyperparameters and deploy them to clients.
Then we sample a starting round $s\sim U[1, T]$, and load the saved model weights $\mathbf{w}_0^{(s)}$ to the global model. 
Starting from this model, we then run the FL algorithm for $T_s \ll T$ 
(e.g. $T_s=1$) rounds to get a new model $\mathbf{w}^{(s+T_s)}$.
Finally, we use the validation performance gain of $\mathbf{w}^{(s+T_s)}$ over $\mathbf{w}^{(s+T_s)}_0$ as reward to update HPN.

RST is more effective to deal with the 
evaluation challenge in pFedHPO than full-fidelity and low-fidelity training methods.
Firstly, since RST only trains FL for $T_s$ rounds at each trail, 
the communication cost of RST is much less than that of the method which runs a complete FL course for evaluation.
In addition, as the starting round of each trial is randomly sampled, in expectation, RST can cover the entire FL course, instead of covering only a few rounds at the beginning of FL course, like low-fidelity methods.
Therefore, by using the RST method, we can achieve better trade-off between the efficiency and accuracy of hyperparameter evaluation for HPN training.

\section{Experiments}
\label{sec:exp}

In the experiments, we first evaluate the effectiveness of our personalized method HPN for tuning client hyperparameters in FL compared to the non-personalized FedHPO methods.
Then we extend the non-personalized FedHPO methods to pFedHPO setting and compare them with HPN to show the advantages of HPN for achieving pFedHPO. 
In addition, we verify the effect of HPN for pFedHPO under different data heterogeneity settings.
Finally, we demonstrate and analyze the convergence of the HPN training process.

\subsection{Setting}

We conduct experiments on NLP (Twitter), graph (Graph-DC), and image (CIFAR-10) datasets.
In our experiments, we tune local hyperparameters for all clients for FedAvg~\cite{deffl} algorithm.
The computation budget for all experiments are 4000 communication rounds for HPO and 400 communication rounds for full-fidelity evaluation, which means that 
all HPO methods consume 4000 communication rounds for their hyperparameter tuning process,
and the full-fidelity FL course for evaluation runs for 400 communication rounds. 
We consider test accuracy the metric of interest and repeat each method 5 times to report the averaged best-seen results.
We implement our method based on FederatedScope \cite{fs}.

\paragraph{Baselines.} 
We compare our method with ten HPO methods. 
Specifically, we consider five traditional HPO methods: random search (RS)~\cite{rs}, Bayesian optimization with a GP model ($\rm{BO}_{GP}$)~\cite{bogp1}, 
Bayesian optimization with a random forest model ($\rm{BO}_{RF}$)~\cite{bogp1}, Hyperband (HB) \cite{hyperband}, and model-based extension of Hyperband
with KDE-based model (BOHB~\cite{bohb}).
In addition, we also consider the combination of FedEx~\cite{fedex} with the above methods by applying these traditional HPO methods as wrappers, leading to five FedHPO methods~\cite{fedhpobench}: RS+FedEx, $\rm {BO}_{RF}$+FedEx, $\rm {BO}_{GP}$+FedEx, HB+FedEx, and BOHB+FedEx.

\paragraph{Twitter} \cite{leaf} is a sentiment analysis dataset, the task of which is to analyze the sentiment of sentences, where each device is a different twitter user.
For the Twitter experiment, we use a subset of 3300 clients. 
The search space consists of local learning rate values of $\{$1e-5, 5e-5, 1e-4, 5e-4, 1e-3, 5e-3, 1e-2$\}$ and weight decay values of $\{$0, 1e-5, 1e-4, 1e-3, 1e-2, 1e-1$\}$.
The local update steps is 10. Following \cite{fedhpobench}, we use the LR model for this experiment.

\paragraph{Graph-DC} \cite{hete-task} is a hetero-task learning dataset. 
There are 13 clients in this dataset, each of which holds a 
molecular graph classification dataset with distinct binary
classes, which means that different clients have different local learning goals. 
For the this experiment, we use a subset of 5 clients.
The search space consisted of local learning rate values of $\{$1e-1, 1e-2, 1e-3, 1e-4$\}$, and local update steps of $\{1,2,3,4\}$. We use the GIN model from \cite{hete-task} for this experiment.

\paragraph{CIFAR-10} \cite{cifar10} is an image dataset which consists of 60,000 color images for image classification.
We split the raw training set to training and validation sets with a ratio $4:1$.
Then we sample 10 subsets for 10 clients according to a Dirichlet distribution with $\alpha$ varying between 0.05 and 5 for different heterogeneity.
The search space consists of local learning rate of $\{$1e-3, 5e-3, 1e-2, 5e-2, 1e-1$\}$, weight decay of $\{$0, 1e-5, 1e-4, 1e-3, 1e-2, 1e-1$\}$, and dropout values of $\{ 0, 0.1, 0.2, 0.3, 0.4, 0.5 \}$. We use the CNN model from~\cite{fedhpobench} for this experiment.

\subsection{Results}

\begin{table}[thbp]\label{table:main}
	\centering
	\begin{tabular}{lccccccccccc}
		\toprule
		Method  &  Graph-DC & Twitter   \\
		\midrule
		RS  & $67.78 \pm 2.14$ & $57.49 \pm 2.98$ \\
		$\rm {BO}_{GP}$  & $68.67 \pm 1.74$ & $57.88 \pm 2.68$   \\ 
		$\rm {BO}_{RF}$  & $68.95 \pm 0.98$ & $57.75 \pm 2.40$  \\
		HB  & $67.93 \pm 2.42$ & $58.22 \pm 1.84$ \\
		BOHB  & $68.55 \pm 2.35$ & $58.10 \pm 2.23$  \\
		RS+FedEx  & $69.36 \pm 2.04$ & $58.85 \pm 1.44$   \\
		$\rm{BO}_{GP}$+FedEx  & $70.07 \pm 1.25$ & $59.46 \pm 1.25$   \\
		$\rm {BO}_{RF}$+FedEx  & $69.90 \pm 1.18$  & $59.33 \pm 1.58$  \\
		HB+FedEx  & $69.52 \pm 1.05$ & $59.83 \pm 1.67$  \\ 
		BOHB+FedEx  & $70.16 \pm 0.88$ & $59.92 \pm 1.93$  \\
		\midrule
		HPN & $\bf{70.47 \pm 1.17}$ & $\bf{60.32\pm 1.65}$\\
		\bottomrule
	\end{tabular}
	\caption{
		Test accuracy ($\%$) of FL model trained with hyperparameters tuned by HPN and non-personalized FedHPO methods on Graph-DC and Twitter datasets. 
		For both datasets, HPN achieves higher test accuracy than all non-personalized FedHPO baselines, indicating that it is more effective to tune client hyperparameters for the same amount of computation budget.  
	}              
\end{table}
\paragraph{Comparison with non-personalized FedHPO methods.}

\begin{table}[thbp]\label{table:pfedhpo}
	\centering
	\begin{tabular}{lccccccccccc}
		\toprule
		Method  &  Test Acc    \\
		\midrule
		RS+FedEx  & $67.36 \pm 3.04$    \\
		$\rm{BO}_{GP}$+FedEx  & $68.07 \pm 3.46$    \\
		$\rm {BO}_{RF}$+FedEx  & $68.90 \pm 2.34$    \\
		HB+FedEx  & $67.88 \pm 3.25$   \\ 
		BOHB+FedEx  & $68.06 \pm 2.88$   \\
		\midrule
		HPN & $\bf{70.47 \pm 1.17}$ \\
		\bottomrule
	\end{tabular}
	\caption{
		Test accuracy ($\%$) of FL model on Graph-DC dataset trained with hyperparameters tuned by HPN and other pFedHPO methods adapted from non-personalized FedHPO methods.
		our method outperforms all the adapted pFedHPO methods by a large
		margin.
		Moreover, the performance of adapted pFedHPO method is even worse than 
		their original non-personalized algorithm,
		indicating that it is difficult
		to directly adapt existing FedHPO methods to pFedHPO
		settings by extending their search space.
	}              
\end{table}

Table~\ref{table:main} shows the results of experiments on Twitter and Graph-DC datasets. 
For both datasets, our proposed method achieves higher weighted average test accuracy than all non-personalized FedHPO baselines.
This indicates that using our personalized FedHPO method to tune client hyperparameters is more effective than all of these non-personalized FedHPO methods for the same amount of computation budget.
This result is also consistent with the observation in previous work~\cite{motley}, which shows that finding a good set of hyperparameters applied to all the clients is difficult due to the data heterogeneity and task heterogeneity in federated learning.


\paragraph{Comparison with personalized FedHPO methods.}
Although our method is the first algorithm for the pFedFBO setting, existing non-personalized FedHPO method can be adapted to the pFedHPO setting by expanding their search space to a Cartesian product of all possible combination of all the clients' local space. 
Take non-personalized FedHPO method RS as example, on Graph-DC dataset, its search 
space consists of local learning rate with 4 different values and local update steps with 4 different values. In this case, there are 16 possible combinations in the search space.
To adapt RS to pFedHPO setting, we can set its search space as the Cartesian product of 5 spaces (for 5 clients), each of which contains 16  groups of hyperparameters. As a result, the whole search space contain $16^5=1,048,576$ combinations.
As this search space is too large to run HPO methods,
we can only sample a subset from this space as search space to run RS. 
Then the finally selected hyperparameters are personalized as each candidate contains different hyperparamsters for different clients.

We adapt five FedHPO methods to pFedHPO setting on Graph-DC dataset. 
As shown in Table~\ref{table:pfedhpo}, HPN outperforms all the adapted methods by a large margin (more than $1.5\%$).
Moreover, we observe that the performance of the adapted method is even worse than that of their original non-personalized method.
The reason is that the search space of these methods for pFedHPO setting is extremely large (more than $1\times 10^6$ combinations) while only a small subset (100 combinations in our experiments) can be sampled.
This indicates that it is difficult to directly adapt existing non-personalized FedHPO methods to pFedHPO settings by directly extending the search space, as this might result in the FedHPO methods not being able to converge within a reasonable computational budget due to the large search space, or lead to the sampled search space being too small to find good hyperparameters in pFedHPO setting.

\begin{table}[t]\label{table:cifar}
	\centering
	\begin{tabular}{l|ccccccccccc}
		\toprule
		Heterogeneity & RS & HPN \\
		\midrule
		$\alpha=0.05$ & $48.94 \pm 0.73$ & $\bf{50.44 \pm 1.26}$ \\
		$\alpha=0.1$ & $61.14 \pm 0.60$ &   $\bf{62.85 \pm  1.47}$ \\ 
		$\alpha=0.5$ & $69.57 \pm 0.44$ &  $\bf{69.77 \pm 0.82}$ \\
		$\alpha=1.0$ & $71.15 \pm 0.32$ & $\bf{71.68 \pm  0.33}$ \\
		$\alpha=5.0$ & $73.67 \pm 0.35$ & $\bf{74.06 \pm 1.02}$\\ 
		\bottomrule
	\end{tabular}
	\caption{
		Test accuracy on CIFAR-10 dataset under different data distribution heterogeneity where a smaller $\alpha$ indicates higher data heterogeneity. Comparing with random search, HPN achieves higher weighted average test accuracy for all data heterogeneity settings. 
		Furthermore, HPN obtains more performance gain for higher data heterogeneity.
	}              
\end{table}

\paragraph{Effectiveness under different data heterogeneity.}
We also evaluate the performance of the proposed method under different data distribution heterogeneity.  
Table~\ref{table:cifar} shows the results from experiments on the CIFAR-10 dataset for different data heterogeneity, with Dirichlet distribution parameter $\alpha$ varying from 0.05 to 5. 
We compared our method against random search. Our proposed method achieved higher weighted average test accuracy for all the data heterogeneity settings. 
Furthermore, our method can achieve more performance gain for higher data heterogeneity.
For instance, 
for a low data heterogeneity setting with $\alpha=5.0$, the test accuracy of our method
is only about $0.4\%$ higher than that of RS,
while for a high data heterogeneity setting with $\alpha=0.05$, our method achieves a test accuracy of $50.44\%$, which is $1.50\%$ higher than that of RS.
These results suggest that our personalized FedHPO method is more effective in searching optimal configurations for different clients in the case of high data heterogeneity.

\begin{figure*}[thbp]
	\label{fig:converge}
	\centering
	\includegraphics[scale=.66]{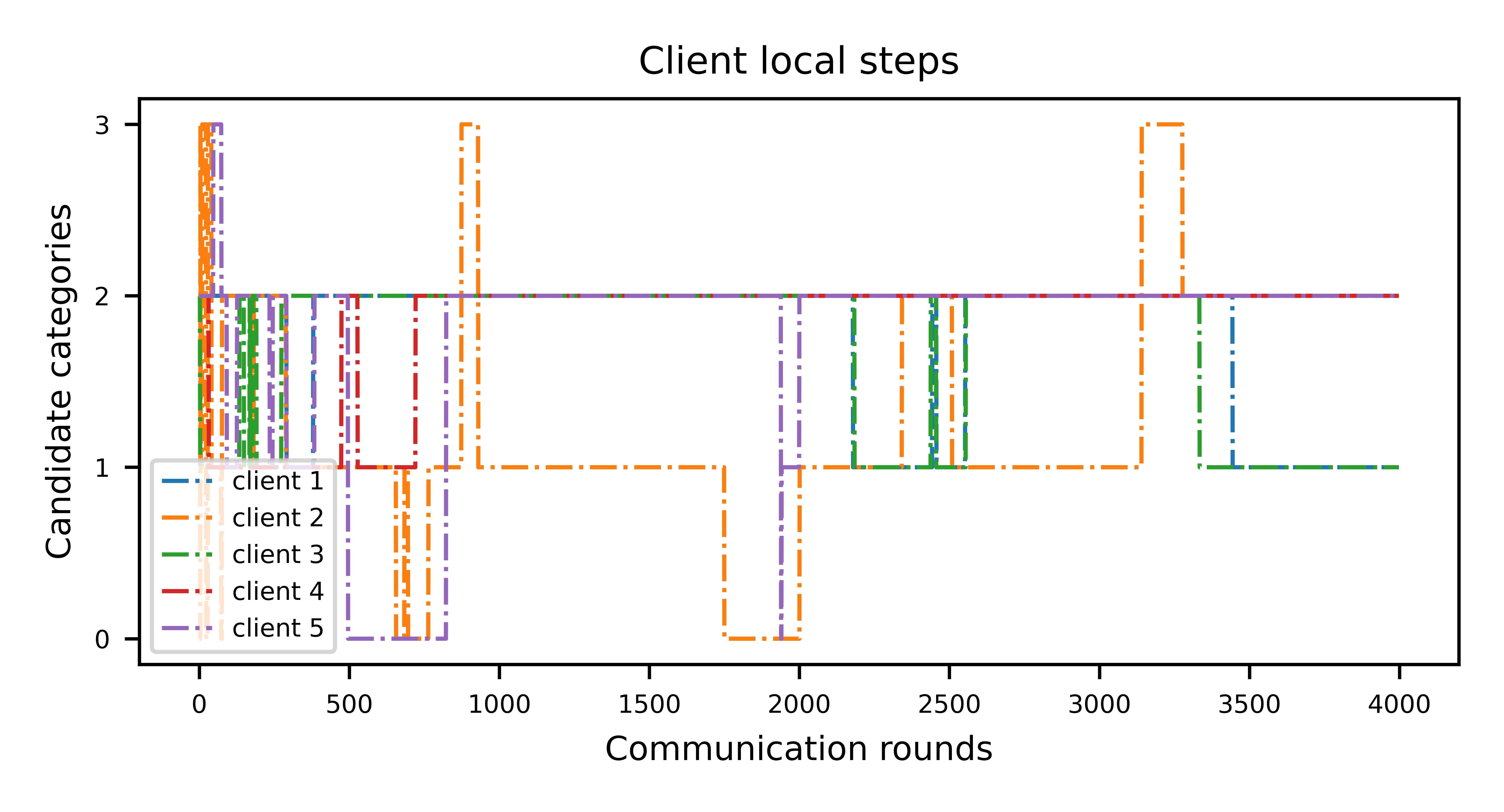}
	\includegraphics[scale=.66]{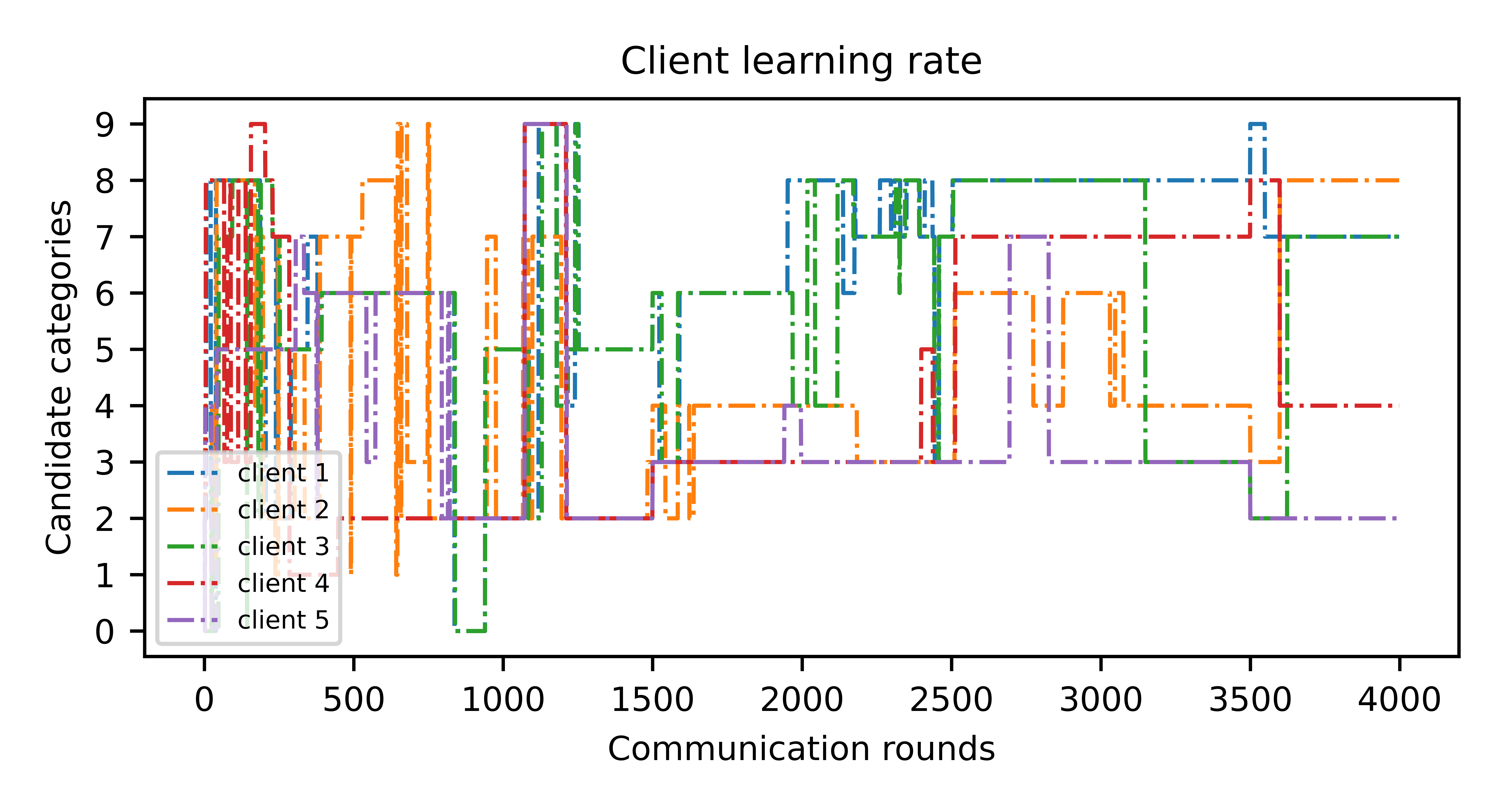}
	\vspace{-0.5cm}
	\caption{    
		The curves about the category corresponding to the maximum logits of HPN output of local steps (top) and local learning rate (bottom) at each communication round during an HPN training process on Graph-DC dataset.
		The output category changes frequently at early learning stage, indicating that HPN tend to explore more possibilities combinations at this stage. 
		At the late training stage, the number of output category changes decreases significantly, indicating that the output of HPN converges to the optimal configurations.
	}
\end{figure*}

\paragraph{Convergence of HPN.}

In order to demonstrate the convergence of our proposed method, we plot curves about the category of the maximum logits of HPN output after each communication round during a HPN training process on Graph-DC dataset.
In Fig.~\ref{fig:converge}, we plot the curves about the output hyperparameters of client local step and local learning rate.
We can observe that at the beginning of the learning stage (the first 1000 communication rounds), the output category changes frequently, indicating that at this stage, HPN is not well-trained and it cannot determine which hyperparameter configurations are superior.
As a result, it tends to explore more possible combinations at this stage.
On the contrary, in the late training period (the last 1000 communication rounds), the number of output category changes decreases significantly.
At this stage, the HPN tends to do more exploitation to the optimal configurations instead of exploring other actions like early stages, which means the HPN becomes certain about its choice of the optimal output, indicating the convergence of the HPN model.

\section{Related Work}
\label{sec:related}
\noindent\textbf{Hyperparameter Optimization (HPO)}.
HPO intends to seek suitable hyperparameter configuration(s) for a considered algorithm to fully unleash its potential.
This procedure is formulated as a optimization problem: $\max_{c\in\mathcal{C}}f(c)$, where the domain of $f$ (i.e., $\mathcal{C}$) corresponds to the search space,
and the value $f(x)$ is often acquired by executing the corresponding algorithm configured by $c$ and reporting the performance.
As $f$ is a black-box function, existing HPO methods essentially solve such a problem in a trial-and-error manner.
The main difference lies in that model-free methods such as random search~\cite{rs} and grid search~\cite{gridsearch} simply explore the search space,
while model-based methods~\cite{bogp1,bogp2,kde} exploit accumulated experience to model the landscape of $f$ and propose their next query.
It is worth noticing that multi-fidelity methods~\cite{hyperband,bohb} have been confirmed to be effective in many cases,
where full-fidelity evaluation of $f(c)$ is time-consuming or even unaffordable.
One of the most practical and popular fidelity dimension adopted by these methods, in making HPO for a training course, is the number of training epochs.
Specifically, they usually conduct a low-fidelity evaluation by executing the training course from scratch or from a snapshot of prior epoch for a few epochs.
Instead, HPN conducts each low-fidelity evaluation, starting from any stage of a training course.

\noindent\textbf{Federated Learning (FL) and FedHPO}.
In general, FL involves multiple participants in a distributed machine learning scenario~\cite{flsurvey,flsurvey1,fedoptimizationsurvey}.
Taking several widely-adopted FL algorithms (e.g., FedAvg~\cite{deffl}, FedOpt~\cite{fedopt}, and FedProx~\cite{fedprox}) for examples,
each training course consists of multiple communication rounds among participants.
When we consider HPO for FL, which is referred to by \subj in this paper, each trial (i.e., a training course) is much more costly than that in a centralized scenario.
To reduce the number of trials \subj methods need, one promising strategy is to make ``concurrent exploration''~\cite{fedex,fedhpobench},
i.e., letting FL participants try different hyperparameter configurations in each communication round.
Moreover, FL imposes another challenge to HPO that the heterogeneity among FL participants~\cite{ditto,fs},
including statistical heterogeneity (e.g., the Non-IIDness among their data) and physical heterogeneity (e.g., various system resources),
might lead to client-wise optimal hyperparameter configurations.
As a result, seeking only one hyperparameter configuration is likely to be sub-optimal.
In this paper, we provide empirical evidence to confirm the existence of such an issue and design HPN to address it.

\section{Conclusions and Future Directions}
\label{sec:con}
In this paper, we first discuss the setting, challenges, and critical factors in the problem-solving of \psubj. Subsequently, we propose several synergistic components to learn HPN for achieving \psubj. To the best of our knowledge, this is the first study of this topic, and extensive empirical studies imply that our proposed HPN constitutes a solid step forward in this area.
Of course, we have not explored some potential issues in \psubj, such as how to ensure the clients honestly report their validation performances, seek hyperparameters that are both competitive and fair for all, and handle adversarial clients.

\balance
\bibliographystyle{icml2021}
\bibliography{ref}

\end{document}